\documentclass[11pt,a4paper]{article}
\usepackage[hyperref]{emnlp-ijcnlp-2019}
\usepackage{times}
\usepackage{latexsym}

\usepackage{amsmath,graphicx}
\usepackage{latexsym}
\usepackage{algorithm}
\usepackage{algpseudocode}
\usepackage{url}
\usepackage{multirow}
\usepackage{CJKutf8} 
\usepackage{textcomp }
\usepackage{comment} 
\aclfinalcopy 

\title{Cross-Lingual Transfer Learning for Question Answering}

\author{Chia-Hsuan Lee\\
National Taiwan University, Taiwan\\
  {\tt r06944037@ntu.edu.tw} \\\And
  Hung-Yi Lee \\
National Taiwan University, Taiwan\\
  {\tt hungyilee@ntu.edu.tw} \\
  }
\date{}

\begin{document}
\maketitle
\begin{abstract}
Deep learning based question answering (QA) on English documents has achieved success because there is a large amount of English training examples.
However, for most languages, training examples for high-quality QA models are not available. 
In this paper, we explore the problem of cross-lingual transfer learning for QA, where a source language  task with plentiful annotations is utilized to improve the performance of a QA model on a target language  task with limited available annotations. 
We examine two different approaches. 
A machine translation (MT) based approach translates the source language into the target language, or vice versa.
Although the MT-based approach brings improvement, it assumes the availability of a sentence-level translation system.
A GAN-based approach incorporates a language discriminator to learn language-universal feature representations, and consequentially transfer knowledge from the source language.
The GAN-based approach rivals the performance of the MT-based approach with fewer linguistic resources. 
Applying both approaches simultaneously  yield the best results.
We use two English benchmark datasets, SQuAD and NewsQA, as source language data, and show significant improvements over a number of established baselines on a Chinese QA task. 
We achieve the new state-of-the-art on the Chinese QA dataset. 
\end{abstract}

\section{Introduction}
Question answering (QA) has drawn much attention in the past few years, and end-to-end deep learning has demonstrated promising performance on QA~\cite{yu2018qanet,xiong2016dynamic,seo2016bidirectional,huang2017fusionnet,wang2018multi,liu2017stochastic,hu2017reinforced,wang2017gated}. 
QA tasks on images~\cite{zitnick2016adopting}, audio~\cite{lee2018odsqa,li2018spoken} or video descriptions~\cite{rohrbach2015dataset} have been studied, but most focus on understanding text documents~\cite{lai2017race,rajpurkar2016squad,clark2018think,joshi2017triviaqa,he2017dureader,nguyen2016ms,trischler2016newsqa}. 

To train end-to-end network-based QA models, a large amount of training examples are needed.
Although English training examples are plentiful, most languages still lack the resources to train high-quality QA models; moreover, annotating training examples for QA is expensive.  
Therefore, it is desirable to transfer the knowledge of a QA model from a source language with many training examples such as English to target languages with fewer training examples. 

Translating the data in source language into target language or vice versa by machine translation (MT) is an intuitive way to achieve cross-lingual transfer learning for QA. In this paper, we first validate the benefits of sentence-level MT-based approaches for transferring QA knowledge; we find that MT-based approaches bring significant improvements.

However, not all the language pairs have high quality sentence-level MT system. We propose using domain adversarial training~\cite{chen2016adversarial} to learn language-invariant latent representations on top of the QA models.  With only word-by-word bilingual dictionary, this approach rivals the performance of the sentence-level MT-based approach. 
Finally, we find that applying both approaches simultaneously yield the best results.

\section{Related Work}
\subsection{Transfer Learning for Question Answering}
Transfer learning is a set of techniques using source domain data to enhance the performance of a model on the target domain.
Transfer learning reduces the required amount of target domain training data;  equivalently, it improves performance in the target domain. 
It has been successfully applied in computer vision~\cite{ganin2016domain} and speech recognition~\cite{shinohara2016adversarial}. 
It is also widely studied on NLP tasks such as sequence tagging~\cite{yang2017transfer}, syntactic parsing~\cite{mcclosky2010automatic}, and named entity recognition~\cite{chiticariu2010domain}.

QA transfer learning has also been studied. 
In probably the first work to apply transfer learning on QA, the authors show that the transferred system is significantly better than a random baseline~\cite{kadlec2016particular}. \citet{min2017question} use source data to pre-train   QA model, and then fine-tune the model with  target data. 
\citet{wiese2017neural} show that it is possible to transfer knowledge from an open domain dataset for improvements on a biomedical dataset. 
\citet{chung2017supervised} study both supervised and unsupervised transfer learning techniques.
\citet{golub2017two} propose a two-stage synthesis network that generates synthetic questions and answers in the target domain without annotations to augment training data. 
However, in all these studies, the source and target domains are in the same language. 
\subsection{Multilingual Question Answering}
Another line of research related to our work is multilingual question answering (MLQA), also called cross lingual question answering (CLQA) in which the machine is required to return an answer in target language with respect to a question in source language using documents in source language \cite{magnini2006overview,sasaki2005overview}. 
One typical approach is using a mono-lingual QA system for the source language and translating resulting answers into the target language \cite{bos2006cross}. Another is translating questions from source language into target language and using a mono-lingual QA system for the target language to return answers \cite{mori2007method}. 
The task in this paper is completely different from MLQA. In this work, the question, document, and answer are all in the same language, and we use training data pairs from another language to improve model performance on target language.

\subsection{Cross-Lingual Transfer Learning}
For knowledge transfer between different languages, it is intuitive to use machine translation (MT) to translate the source language into the target language, or vice versa. 
Using MT to achieve knowledge transfer between different languages has been studied on sentiment analysis~\cite{mohammad2016translation}, spoken language understanding~\cite{stepanov2013language,schuster2018cross} and question answer~\cite{asai2018multilingual}. It is also possible to train language model to obtain cross-lingual text representations and further improve QA performance in different languages using parallel data~\cite{lample2019cross,mulcaire2019polyglot} or even without parallel data~\cite{devlin2018bert}\footnote{We refer to the multilingual version in \url{https://github.com/google-research/bert/blob/master/multilingual.md}}. 

The performance of MT-based approaches depends highly on the quality of the available MT systems. 
Due to the lack of reliable MT tools for some language pairs, approaches that require only limited linguistic knowledge resources between the source and the target languages have been proposed. 
\citet{zirikly2015cross} only assume the availability of a small bilingual dictionary and train the same model to tag the target corpus.
\citet{lu2011joint} augment the available labeled data in the target language with unlabeled parallel data. 

Compared with the above tasks, cross-lingual transfer learning for QA without a sentence-level MT system is even more challenging because returning the right answer requires that the machine comprehends the document and conducts cross-sentence reasoning. 
By considering the source and target languages to be two different domains, we propose using domain adversarial training~\cite{chen2016adversarial,kim2017cross} to cause the QA model to learn language-invariant latent representations, which encourages knowledge transfer between languages.

\section{Task Description}

Of the many existing QA settings, here we focus on extraction-based QA (EQA), but it is possible to adopt the approaches described in this study to other types of QA tasks.  
In EQA, each example is a triple $(q, d, a)$ in which $q$ is a question and $d$ is a multi-sentence document that contains the answer $a$. 

We seek to improve EQA model performance on the target EQA dataset by transferring knowledge from the source EQA dataset.
The target and source datasets are in different languages. 
For training, we have a large amount of training examples from the source domain, but only a few examples from the target domain. 
The testing data is in the target language.
Due to the limited number of training examples in the target domain, training a QA model from the target domain training set and then applying the model on the testing set of the target domain would yield poor performance. 
The goal of this research is to use the training examples from the source domain to improve performance on the testing set of target domain.

In this paper, the target language is Chinese and the source language English, but the following discussion can be applied to other language pairs.
Compared with other language pairs, knowledge transfer between English and Chinese is difficult because there are no common characters in the two languages.
As such we cannot use shared character embeddings~\cite{yang2017transfer}. 
\section{Cross-Lingual Transfer Learning for QA}

In this section, we present the cross-lingual transfer learning techniques for QA. 
In the first approach, we use an MT system to translate the sentences from one language to the other.
In the second approach, language-invariant representations are learned, and only a word-by-word bilingual dictionary is needed. 

\subsection{Machine Translation Based Approaches}
There are two ways to use an (sentence level) MT system.
\begin{itemize}
\item \textit{Train-on-Target}~\cite{garcia2012combining}: 
In this approach the MT system is used to translate the documents, questions ans answers of the source domain into the target domain. As the source domain training examples are  translated into the target domain, they can be used to train the QA model for the target language. 
\item \textit{Test-on-Source}~\cite{he2013multi}: 
In this approach the MT system is used to translate the documents, questions and answers of the target domain into the source domain, so the QA model trained on the source language can be directly applied on the translated target domain testing data. ~\cite{asai2018multilingual} proposed to back-translate the answer into target language using white-box Neural Machine Translation System while ours utilized off-the-shelf MT system (e.g. Google Translate). In our experiments, this approach leads to inferior performance to \textit{Train-on-Target}. See supplemental material for details. 
\end{itemize}
In both approaches, we split the document into sentences and translate them individually. The translated document is obtained by concatenating all translated sentences. We remove the data examples in which answer spans are no longer recoverable in the documents after translation. 

\subsection{GAN-based Approach}
\label{GAN-intro}
In this approach, we propose a QA model which projects the sentences of two different language domains onto the same  space, so the model benefits from  utilizing training examples from both source and target languages.
Instead of utilizing a sentence-level MT system, this approach requires only a word-by-word bilingual dictionary.
 
\textbf{Bilingual Embeddings}.
First of all, we seek to produce word embeddings in the two languages in the same continuous space. 
Aligning word embeddings between two languages without supervision is an active research field~\cite{conneau2017word}; this approach yielded poor performance in preliminary experiments. 
We conjecture that this is due to fundamental linguistic differences between the source language (English) and the target language (Chinese); thus the embedding space could not be properly aligned\footnote{It has been shown that aligning word embeddings between English and Chinese is more challenging than for other language pairs~\cite{conneau2017word}.}.

Here we use a word-by-word bilingual dictionary to solve the problem. 
For each word $w$ in the source language, we utilize the dictionary to fetch the corresponding word $w'$ in the target language. 
Then we use the word embedding of word $w'$ in the target language as the embedding of word $w$ in the source language. Thus, as the two languages actually use the same set of word embeddings,
the model is more likely to bootstrap knowledge from both languages during training and eventually improve performance when testing on the target language. 

\begin{figure*}[t]
  \centering
  \includegraphics[width=150mm]{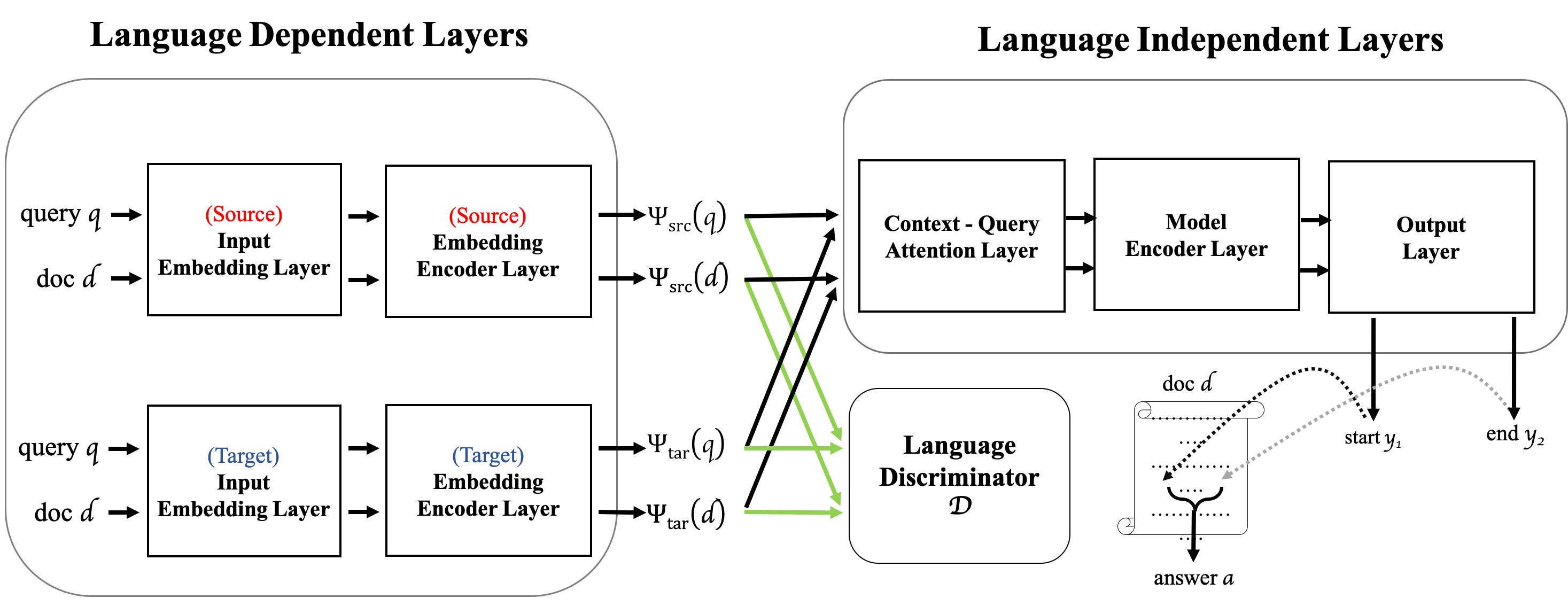}
  \caption{Architecture of proposed QA model and language discriminator.  Each language uses the same network architecture for the language-independent layers.}
  \label{fig:overview}
\end{figure*}

\textbf{Language-Dependent and -Independent Layers}.
However, in preliminary experiments, we found that using the same set of word embeddings is not sufficient to achieve efficient joint training.
This is because the syntactic structures of the two languages are often very different. 
To address this problem, we propose the QA model architecture in Fig.~\ref{fig:overview}. 
This is a general concept, and is independent of the network architecture. 

The QA model in Fig.~\ref{fig:overview} separately encodes sentences from the target and source domains with two sets of \textit{language-dependent layers} to handle the different underlying linguistic structures. 
Each language has its own language-dependent layers, each of which takes a sequence of words of a language as input and outputs a sequence of vectors. 
The length of the output vectors is the same as the length of the input word sequence.
The input word sequence can be a query $q$ or document $d$.
We use $\Psi_\text{tgt}(q)$ and $\Psi_\text{tgt}(d)$ ($\Psi_\text{src}(q)$ and $\Psi_\text{src}(d)$) to represent the output vector sequence given query $q$ and document $d$ in the target (source) language respectively. 

The subsequent layers in QA models are \textit{language-independent layers} which take the $\Psi_\text{tgt}(q)$ and $\Psi_\text{tgt}(d)$ ($\Psi_\text{src}(q)$ and $\Psi_\text{src}(d)$) as input, and output the final answer. 
The parameters of language-independent layers are shared across different languages.

\textbf{Adversarial Training}.
Although it may be that the two language-dependent layers occasionally learn common latent representations, adversarial learning can further 
encourage the outputs of the language-dependent layers to be language-invariant.   
We incorporate a language discriminator $D$ whose job is to identify the language of the input vector sequence from the output of $\Psi_\text{tgt}$ and $\Psi_\text{src}$. $D$ is also shown in Fig.~\ref{fig:overview}. 
The intuition is that if the output of $\Psi_\text{src}$ is indistinguishable from $\Psi_\text{tgt}$, the features they extract are language-invariant, making it easier for the following layers of the QA model to use the knowledge from the training examples of both source and target languages. 

The discriminator $D$ is learned to minimize $L_\text{dis}$ below.
\begin{equation}
\begin{split}
L_\text{dis}  = E_{(q,d,a) \sim \text{target}}  \,[\log D(\Psi_\text{tgt}(q)) \\ +\log D(\Psi_\text{tgt}(d))] \\
+ E_{(q,d,a) \sim \text{source}} \, [\log (1 -  D(\Psi_\text{src}(q)) \\ +\log (1 -  D(\Psi_\text{src}(d))], 
\end{split}
\label{eq2}
\end{equation}
where $D$ is the language discriminator. 
In (\ref{eq2}), given a training example from the target language ($(q,d,a) \sim \text{target}$), $D$ learns to assign a lower score to the $q$ and $d$ in that example, that is, minimize $D(\Psi_\text{tgt}(q))$ and $D(\Psi_\text{tgt}(d))$.  
Conversely, given a training example from the source language ($(q,d,a) \sim \text{source}$), $D$ learns to assign larger values to $q$ and $d$.

The two language-dependent layers in the QA model are trained to maximize  $L_\text{dis}$ while minimizing the loss for QA, $L_\text{qa}$.
The loss function $L_\text{pri}$ for language-dependent layers is
\begin{equation}
    L_\text{pri} =  L_\text{qa} - \lambda_G L_\text{dis}, \label{eq:L_pri}
\end{equation}
where $\lambda_G$ is a hyperparameter.
Because the parameters of language-independent layers are independent of the loss of the language discriminator, the loss function of language-independent layers $L_\text{pub}$ is equivalent to $L_\text{qa}$, that is, $L_\text{pub} =  L_\text{qa}$.  

The definition of $L_\text{qa}$ depends on the specific QA task.
In EQA, each example is labeled with a span in the document containing the answer. 
The QA model predicts the probability of each position in the document being the start or end of an answer span. 
The QA loss function $L_{qa}$ is defined as the negative sum of the log probabilities of the predicted distributions indexed by true start and end indices, averaged over all the training examples: 
\begin{equation}
\begin{split}
  L_{qa} = -\frac{1}{N} \sum_{i}^{N} [\log(p^{1}_{y^{1}_i})+\log(p^{2}_{y^{2}_i})]
\end{split}
\label{lqa}
\end{equation}
where $N$ is the number of training examples and $y_1^i$ and $y_2^i$ are respectively the ground-truth starting and ending positions of example $i$. 
$p^1$ and $p^2$ are respectively the probabilities of the starting and ending positions. 

The QA model and the language discriminator are learned in an adversarial manner.
The training procedure of our model is summarized in Algorithm~\ref{alg:Framwork}.

\begin{algorithm*}[htb]
  \caption{Training procedure.\\
$k$ is the number of learning steps for the discriminator $D$ in each iteration. $\lambda_{G}$ is the discriminator loss coefficient which is initially zero and scales up with the training steps.
$B$ is the batch size.
Details on these hyperparameters are provided in the Experiments section.}
  \label{alg:Framwork}
  \begin{algorithmic}[1]
    \Require
    
    \For{$i = 0,\ldots,\mathit{num\_iter}$}
    \For{$k$ steps}
       \State $ {(q^i_t,d^i_t,a^i_t)}_{i=1}^B \leftarrow$ Sample a batch of training examples from the target domain 
       \State $ {(q^i_s,d^i_s,a^i_s)}_{i=1}^B \leftarrow$ Sample a batch of training examples from the source domain
       \State Update discriminator $D$ by minimizing its loss function:
\State $L_\text{dis}  = \sum_{i=1}^B  \,[\log D(\Psi_\text{tgt}(q^i_t))+\log D(\Psi_\text{tgt}(d^i_t))] 
+ \sum_{i=1}^B \, [\log (1 -  D(\Psi_\text{src}(q^i_s))+\log (1 -  D(\Psi_\text{src}(d^i_s))]$ 
 \EndFor
 
 \State Update language-dependent layers by minimizing their loss function:
 \State   $L_\text{pri} =  L_\text{qa} - \lambda_G L_\text{dis}$  ($L_\text{qa}$ defined in (\ref{lqa}) )
  \State Update language-independent layers by minimizing their loss function:    \State   $L_\text{pub} =  L_\text{qa}$

    \EndFor
  \end{algorithmic}
\end{algorithm*}

\section{Experiments}

\subsection{Question Answering Model} 
Of the numerous models proposed for EQA, we chose QANet~\cite{yu2018qanet} as the base model due to its good performance; also, thanks to its lack of recurrent layers, it can be trained efficiently. 
It is possible to replace QANet with other QA models as long as they can be separated into language-dependent and -independent layers.
Below we briefly introduce the architecture of QANet. 
The network architecture can be found in Fig.~\ref{fig:overview}.
Note that the concatenation of the language-dependent and -independent layers forms the original QANet.

In the language-dependent layers, the \textit{Input Embedding Layer} first produces the word and character embedding for each word in $q$ or $d$, 
after which an \textit{Embedding Encoder Layer}, which consists of an encoder block, is used to model the temporal interactions between words and refine them to contextualized representations. 
The encoder block is composed exclusively of depthwise separable convolutions and self-attention.
The intuition is that convolution components model local interactions and self-attention components model global interactions. 
We found it is necessary to include the embedding encoder layer in the language-dependent layers,
because in preliminary experiments, with only the input embedding layer, the language-dependent layers do not generate language-invariant representation. 

In language-independent layers, a \textit{context-query attention layer} generates the question-document similarity matrix and computes the question-aware vector representations of the context words. 
Likewise, we attempted putting the context-query attention layer in the language-dependent layers instead of the language-independent layers, but it makes the discriminator too easy to discriminate and leads to degrading performance. 
After the context-query attention layer, a \textit{model encoder layer} consisting of seven encoder blocks captures the interaction among the context words conditioned on the question. 
Finally, the \textit{output layer} predicts the start position and end position in the document and then extracts the answer from the document.

\begin{center}
\begin{table}[]
\centering
\caption{An example (document $d$, query $q$, and answer $a$) in DRCD with its english translation.} 
\begin{CJK}{UTF8}{bkai}{}
\label{tab:EXAMPLE}
\begin{tabular}{|c || p{6cm} | }
\hline
\multirow{5}{*}{$d$} & 
 “...所有南亞的主要書寫系統事實上都用於梵語文稿的抄寫。自19世紀晚期，天城文被定為梵語的標準書寫系統...” 
  “...all the major writing systems in South Asia are actually used for the transcription of Sanskrit manuscripts. Since the late 19th century, Tianchengwen has been designated as the standard writing system for Sanskrit...”\\
\hline
\multirow{3}{*}{$q$}  & 天城文在何時成為梵語的標準書寫系統？ 
When did Tianchengwen become the standard writing system for Sanskrit? \\
\hline
\multirow{1}{*}{$a$} &  19世紀晚期 Late 19th century\\
\hline
\end{tabular}
\end{CJK}
\end{table}
\end{center}

\subsection{Corpora} \label{sec:corpus}
We use the SQuAD English corpus, the NewsQA English corpus and the DRCD Chinese corpus.
These three corpora are introduced as below. 

\textbf{Source Data -- SQuAD}.
For the source-language dataset, we chose the SQuAD~\cite{rajpurkar2016squad} training set, in which documents are a set of Wikipedia articles, and the answer with respect to the given question is always a span in the document. This training set contains 87,500 question-answer pairs. 
The average article length is 250 words, and the average question length is 10 words. 

\textbf{Source Data -- NewsQA}.
In order to test the generality of our proposed approach, we conduct experiments on another source-language dataset, NewsQA~\cite{trischler2016newsqa} training set, in which documents are a set of CNN news articles. 
The answer with respect to the given question is also a span in the document. In original NewsQA, there are unanswerable questions specifically designed to test the reasoning ability of model. We remove these questions and leave the challenge of identifying the unanswerable questions for future work. 
To eliminate the difference in article length as a possible cause of trivial discrimination between source and target data, we remove the articles with lengths longer than 600 words.
The resulting training set contains 42,300 question-answer pairs. 
The average article length is 378 words, and the average question length is 6 words.

\textbf{Target Data -- DRCD}.
For the target language dataset, we used the Delta Reading Comprehension Dataset (DRCD)~\cite{shao2018drcd}, 
an EQA dataset in which each document is a set of Chinese Wikipedia articles, and the answer with respect to the given question is always a span in the document, as in SQuAD. 
An example of DRCD and its English translation is shown in Table~\ref{tab:EXAMPLE} respectively. 
In DRCD, the training set contains 26,936 questions with 8,014 paragraphs, which is smaller than the SQuAD training set; the testing set contains 3,524 questions with 1,000 paragraphs\footnote{The testing set mentioned here is actually a development set. The real testing set is not publicly available yet.}. 
After word segmentation, the average number of words per document is 262, and the average number of words per question is 13, which are both slightly more than the SQuAD counterparts.
In DRCD every question has only one ground-truth answer, as opposed to SQuAD's three, which makes it more difficult for the model prediction to match the ground-truth. 
\subsection{Experimental Setup}
The word embeddings in all our experiments were initialized from the 300-dimension pre-trained Fasttext embeddings~\cite{bojanowski2016enriching} and fixed during training for both English and Chinese. The word-by-word bilingual dictionary used for naive word-by-word translation was provided by Google Machine Translation.
  We used JIEBA\footnote{Python library JIEBA: \url{https://pypi.org/project/jieba/}} to segment Chinese sentences into words. The resulting word vocabulary size for DRCD was around 110,000. 
  We used Google Machine Translation for the MT-based approaches\footnote{Obtained from \url{https://cloud.google.com/translate/} in November, 2018 }.
  
\subsection{Training details}
In the QANet embedding encoder layer, the convolutional blocks and self-attention blocks encode the words from both document and question into contextualized word representations. We chose to incorporate our language discriminator on top of this layer to encourage more explicit alignment between feature representations.

We adopted the discriminator design from~\cite{gulrajani2017improved} for our language discriminator: it stacks five residual blocks of 1D convolutional layers with 96 filters and a filter size of 5 followed by one linear layer to convert each input vector sequence into a scalar value.
All models used in the experiments were trained with a batch size of 24, using the Adam optimizer with learning rate of 1e-3 until convergence. 
We adopted the architecture suggested by the QANet paper, but as we found that using the suggested hyperparameters did not yield optimal results, we set the hidden state size to 96 across all layers and the number of self-attention heads to 2. We also used L2 norm regularization, gradient clipping, and moving averages of all weights with an exponential decay rate of 0.999.

We used a variable weight for the discriminator loss
coefficient $\lambda_{G}$. We initially set $\lambda_{G}$ to 0, training the whole model like a normal QA model. 
Then we slowly increased $\lambda_{G}$ to 0.001 over the first 30000 training steps to slowly encourage the model to produce invariant representations. 
Without this scheduling technique, we observed that the model was overly influenced by the disciminator loss and hindered from performing the normal QA task.
$k$ in Algorithm~\ref{alg:Framwork} was set to 5.

\subsection{Evaluation Measures}
The evaluation metrics we used were exact match (EM) and the (macro-averaged) F1 score.
If the predicted text answer and the ground-truth text answer are exactly the same, then the EM score is 1, and 0 otherwise.
The F1 score is based on the precision and recall. 
Precision is the percentage of Chinese characters in the predicted answer that occur in the ground-truth answer, and recall is the percentage of Chinese characters in the ground-truth answer that also occur in the predicted answer. 
The EM and F1 scores of each testing example were averaged as the final EM and F1 score. 
We removed all the punctuation in the answers and used the standard evaluation script from SQuAD~\cite{rajpurkar2016squad} to evaluate the performance. 
\subsection{MT-based Approach}
The train-on-target results and results of competing approaches are shown in Table~\ref{tab:MT}. We translate the sentences of SQuAD and NewsQA into Chinese using Google Machine Translation System. The resulting corpora are denoted  as SQuAD (MT) and NewsQA (MT). We also translate SQuAD and NewsQA into Chinese using only word-by-word bilingual dictionary. The results are denoted as SQuAD (word-by-word) and NewsQA (word-by-word).
Rows (a) to (e) are the established baselines when these models are directly trained on the DRCD training set. 
Row (f) is the results of human performance. 
Row(g) is the result when QANet is trained on untranslated SQuAD, and row (h) is the results for jointly training on both SQuAD and DRCD\footnote{The subsets of data are combined and shuffled before jointly training}. 
Row (k) is the results when QANet is trained only on SQuAD (MT), and row (l) is the results for jointly training on both SQuAD (MT) and DRCD. 
Row (o) is the results when QANet is trained only on SQuAD (word-by-word), while row (p) is the results for jointly training using both SQuAD (word-by-word) and DRCD. Rows (i), (j), (m), (n), (q) and (r) are the results for NewsQA. We see that adding more English training data without translation yields no improvements. (row (h) and (j) v.s. row (e)
We also see that training with additional data from different corpora together yields improvement. Finally, we find that the performances of word-by-word translation are inferior to those of machine translation.

\begin{table}[]
\centering
\caption{EM/F1 train-on-target scores over DRCD. FusionNet is denoted as F-Net.}
\label{tab:MT}
\begin{tabular}{|c|c|c|c|c|}
\hline
\multicolumn{2}{|c|}{{\textbf{Baselines}}} & \multicolumn{1}{|c|}{} & EM & F1   \\
\hline
\hline
\multicolumn{2}{|c|}{{\cite{shao2018drcd}}} & (a) & - & 53.78 \\
\multicolumn{2}{|c|}{{BiDAF \cite{seo2016bidirectional}}} & (b) & 56.45 & 70.57 \\
\multicolumn{2}{|c|}{{DrQA \cite{chen2017reading}}} & (c) & 63.21 & 74.11 \\
\multicolumn{2}{|c|}{{F-Net \cite{huang2017fusionnet}}} & (d) & 57.54 & 70.86 \\
\multicolumn{2}{|c|}{{QANet (Baseline)}} & (e)   & 66.10 & 78.01 \\
\multicolumn{2}{|c|}{{Human Performance}} & (f)  & 80.43 & 93.30 \\
\hline\hline
\multicolumn{2}{|c|}{\textbf{QANet}} & \multicolumn{1}{|c|}{} & EM & F1  \\
\hline
\multicolumn{2}{|c|}{{\emph{SQuAD}}} & (g)  & 3.00 & 12.65 \\
\multicolumn{2}{|c|}{+DRCD} & (h)  & 66.56 & 78.67 \\
\multicolumn{2}{|c|}{{\emph{NewsQA}}} & (i)  & 0.93 & 7.51 \\
\multicolumn{2}{|c|}{+DRCD} & (j)  & 66.23 & 78.45 \\
\hline
\multicolumn{2}{|c|}{{\emph{SQuAD (MT)}}} & (k)  & 53.50 & 72.22 \\
\multicolumn{2}{|c|}{{+DRCD}} & (l)  & 74.20 & 85.67 \\
\multicolumn{2}{|c|}{\emph{{NewsQA (MT)}}} & (m) & 22.42 & 35.99 \\
\multicolumn{2}{|c|}{{+DRCD}} & (n) & 68.98 & 81.41 \\
\hline\hline
\multicolumn{2}{|c|}{{\emph{SQuAD (word-by-word)}}} & (o)  & 18.89 & 37.98 \\
\multicolumn{2}{|c|}{{+DRCD}} & (p)  & 68.89 & 81.31 \\
\multicolumn{2}{|c|}{\emph{{NewsQA (word-by-word)}}} & (q) & 11.74 & 26.82 \\
\multicolumn{2}{|c|}{{+DRCD}} & (r) & 67.16 & 79.52 \\
\hline
\end{tabular}
\end{table}

\begin{figure}[t]
  \centering
  \includegraphics[width=\linewidth]{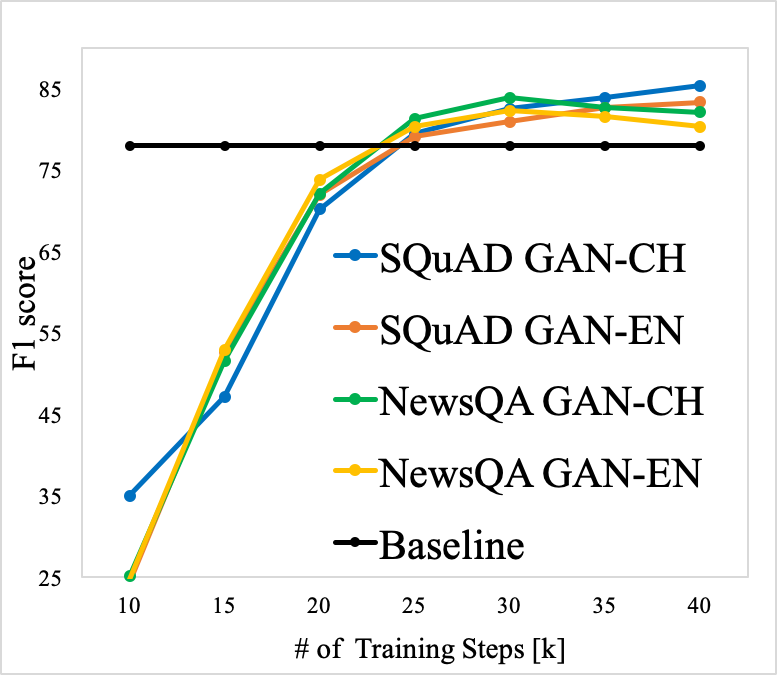}
  \caption{Results of proposed GAN-based approaches on DRCD testing set}
  \vspace{-4mm}
  \label{fig:curve_full}
\end{figure}

\begin{figure}[thb]
  \centering
  \includegraphics[width=\linewidth]{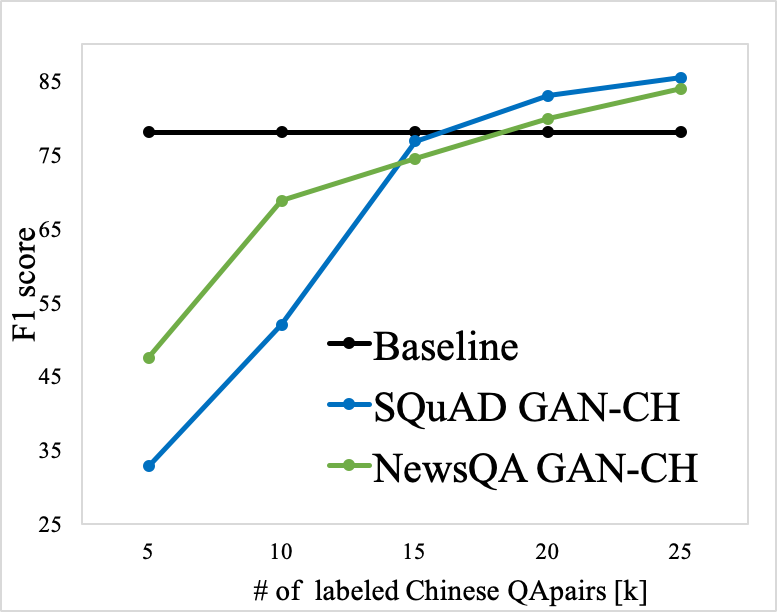}
  \caption{Performance curves given labeled training data from target domain. Vertical axis shows F1 score. Performance curve of GAN-CH is plotted to compare with baseline model.}
  \label{fig:label_filter}
\end{figure}
\subsection{GAN-based Approach}
Here we show the results of the experiments on the GAN-based approach in which two language-dependent layers and a language discriminator are included in the QA model. 
It is possible to update $\Psi_\text{src}$ and $\Psi_\text{tgt}$ simultaneously to minimize $L_\text{pri}$ in (\ref{eq:L_pri}), but it leads to unstable training. 
Therefore, in real implementations, we update either $\Psi_\text{src}$ or $\Psi_\text{tgt}$ to minimize $L_\text{pri}$ in (\ref{eq:L_pri}); the other only  minimizes $L_\text{qa}$ in (\ref{eq:L_pri}) and ignores $L_\text{dis}$.
The above training strategy results in more stable training.
When $\Psi_\text{src}$ minimizes $L_\text{pri}$ while $\Psi_\text{tgt}$ minimizes $L_\text{qa}$, we are steering the output of $\Psi_\text{src}$ toward $\Psi_\text{tgt}$. 
The results are reported in Table~\ref{tab:GAN}. 
Row (a) is the results from row (e) in Table~\ref{tab:MT}. 
Rows (b) to (d) are the results using SQuAD as source data. 
We find that even without the discriminator, the QA model with language-dependent layers brings improvement (row (b) vs. row (p) in Table~\ref{tab:MT}).

The language discriminator $D$ is used in rows (c) and (d).
GAN-CH and GAN-EN represent the results when updating $\Psi_\text{src}$ or $\Psi_\text{tgt}$  to minimize $L_\text{pri}$   respectively. Incorporating adversarial learning is helpful (rows (c) and (d) vs. row (b)). 
We see that the performance of row (c) is better than row (d), attesting the greater benefit of aligning source domain representations to target domain representations. 

The GAN-CH approach rivals the performance of the best setting in MT-based approaches (row (c) vs. row (l) in Table~\ref{tab:MT}).
Note that GAN-based approaches in row (c) utilize only a word-by-word bilingual dictionary and do not utilize any sentence-level MT tool. 
Also, it is beneficial to apply our proposed language-dependent layers and language discriminator as we obtain improvements over base model with the same resources (row (d) vs. row (p) in Table~\ref{tab:MT}).
Rows (e) to (g) are the results using NewsQA as source data.
Similar trend is found on NewsQA.

In Figure~\ref{fig:curve_full}, we also compare the training process between rows (c), (d), (f) and (g). 
We observe testing set F1 scores of GAN-EN that grow quickly but end up inferior to that of GAN-CH. 
This could indicate early convergence. 

Next, in Figure~\ref{fig:label_filter}, we show the performance curves of GAN-CH with various amounts of labeled Chinese (target) data. 
We see that the proposed approach requires only 15,000 Chinese QA pairs to rival the baseline model trained with 25,000 QA pairs; this demonstrates the effects of cross-lingual transfer learning.

\subsection{Combining MT and GAN approaches}
Instead of using word-by-word translation and bilingual embeddings to encode source language data in section~\ref{GAN-intro}, one can use MT system to translate source language data into target language before encoding the results using target language word embeddings. The refined source language data and target language data are then fed into the model in Fig.~\ref{fig:overview}.
We found that using both MT-based approaches and GAN yielded the best performance. The results are shown in Table~\ref{tab:MT+GAN}.

\begin{table}
\centering
\caption{EM/F1 scores of GAN-based approaches on DRCD. 
\textit{Dependent} denotes QA model trained without discriminator;
\textit{MT} denotes the  MT-based approach;
\textit{GAN-CH} denotes the output of the English dependent layer steered toward Chinese;
\textit{GAN-EN} denotes the output of the Chinese dependent layer steered toward English; \textit{MT+GAN-CH} denotes the combination of \textit{MT} and \textit{GAN-CH}.}
\label{tab:GAN}
\begin{tabular}{|c|c|c|c|c|}
\hline
\multicolumn{2}{|c|}{{Approaches}} & \multicolumn{1}{|c|}{} & EM & F1 \\
\hline
\hline
\multicolumn{2}{|c|}{{Baseline}} & (a) & 66.10 & 78.01 \\
\hline\hline
\multicolumn{5}{|c|}{{\emph{SQuAD}}}\\
\hline
\multicolumn{2}{|c|}{{Dependent}} & (b)& 70.97 & 81.92 \\
\multicolumn{2}{|c|}{{+GAN-CH}} & (c)  & 74.00 &  85.35\\
\multicolumn{2}{|c|}{{+GAN-EN}} & (d)  & 70.97 & 83.36 \\
\hline\hline
\multicolumn{5}{|c|}{{\emph{NewsQA}}}\\
\hline
\multicolumn{2}{|c|}{{Dependent}} & (e) & 67.96 & 80.51 \\
\multicolumn{2}{|c|}{{+GAN-CH}} & (f) & 71.73 & 83.90\\
\multicolumn{2}{|c|}{{+GAN-EN}} & (g) & 69.25 & 82.28 \\
\hline
\end{tabular}
\end{table}

\begin{table}
\centering
\caption{EM/F1 scores of combining MT and GAN-based approaches on DRCD. }
\label{tab:MT+GAN}
\begin{tabular}{|c|c|c|c|c|}
\hline
\multicolumn{2}{|c|}{{Approaches}} & \multicolumn{1}{|c|}{} & EM & F1 \\
\hline\hline
\multicolumn{5}{|c|}{{\emph{SQuAD}}}\\
\hline
\multicolumn{2}{|c|}{{MT+GAN-CH}} & (a)  & 75.12 & 87.26 \\
\hline\hline
\multicolumn{5}{|c|}{{\emph{NewsQA}}}\\
\hline
\multicolumn{2}{|c|}{{MT+GAN-CH}} & (b) & 72.79 & 84.94 \\
\hline
\end{tabular}
\end{table}

\section{Conclusion}
We investigate several cross-lingual transfer learning approaches for QA.
First, we apply sentence-level MT-based approaches, which bring significant improvements over target-domain testing data. 
Second, by incorporating domain adversarial learning, the GAN-based approach learns language-invariant latent representations and consequentially transfers knowledge from the source domain. 
Given only a word-by-word bilingual dictionary, a GAN-based approach rivals the performance of the best MT-based approach, and integrating MT-based and GAN-based approaches yields the best results. We conducted experiments using SQuAD and NewsQA as source language datasets and achieved new state-of-the-art on a Chinese QA dataset: DRCD.

\bibliography{emnlp-ijcnlp-2019}

\begin{thebibliography}{51}
\expandafter\ifx\csname natexlab\endcsname\relax\def\natexlab#1{#1}\fi

\bibitem[{Asai et~al.(2018)Asai, Eriguchi, Hashimoto, and
  Tsuruoka}]{asai2018multilingual}
Akari Asai, Akiko Eriguchi, Kazuma Hashimoto, and Yoshimasa Tsuruoka. 2018.
\newblock Multilingual extractive reading comprehension by runtime machine
  translation.
\newblock \emph{arXiv preprint arXiv:1809.03275}.

\bibitem[{Bojanowski et~al.(2016)Bojanowski, Grave, Joulin, and
  Mikolov}]{bojanowski2016enriching}
Piotr Bojanowski, Edouard Grave, Armand Joulin, and Tomas Mikolov. 2016.
\newblock Enriching word vectors with subword information.
\newblock \emph{arXiv preprint arXiv:1607.04606}.

\bibitem[{Bos and Nissim(2006)}]{bos2006cross}
Johan Bos and Malvina Nissim. 2006.
\newblock Cross-lingual question answering by answer translation.
\newblock In \emph{CLEF (Working Notes)}.

\bibitem[{Chen et~al.(2017)Chen, Fisch, Weston, and Bordes}]{chen2017reading}
Danqi Chen, Adam Fisch, Jason Weston, and Antoine Bordes. 2017.
\newblock Reading {Wikipedia} to answer open-domain questions.
\newblock In \emph{Association for Computational Linguistics (ACL)}.

\bibitem[{Chen et~al.(2016)Chen, Sun, Athiwaratkun, Cardie, and
  Weinberger}]{chen2016adversarial}
Xilun Chen, Yu~Sun, Ben Athiwaratkun, Claire Cardie, and Kilian Weinberger.
  2016.
\newblock Adversarial deep averaging networks for cross-lingual sentiment
  classification.
\newblock \emph{arXiv preprint arXiv:1606.01614}.

\bibitem[{Chiticariu et~al.(2010)Chiticariu, Krishnamurthy, Li, Reiss, and
  Vaithyanathan}]{chiticariu2010domain}
Laura Chiticariu, Rajasekar Krishnamurthy, Yunyao Li, Frederick Reiss, and
  Shivakumar Vaithyanathan. 2010.
\newblock Domain adaptation of rule-based annotators for named-entity
  recognition tasks.
\newblock In \emph{Proceedings of the 2010 conference on empirical methods in
  natural language processing}, pages 1002--1012. Association for Computational
  Linguistics.

\bibitem[{Chung et~al.(2017)Chung, Lee, and Glass}]{chung2017supervised}
Yu-An Chung, Hung-Yi Lee, and James Glass. 2017.
\newblock Supervised and unsupervised transfer learning for question answering.
\newblock \emph{arXiv preprint arXiv:1711.05345}.

\bibitem[{Clark et~al.(2018)Clark, Cowhey, Etzioni, Khot, Sabharwal, Schoenick,
  and Tafjord}]{clark2018think}
Peter Clark, Isaac Cowhey, Oren Etzioni, Tushar Khot, Ashish Sabharwal, Carissa
  Schoenick, and Oyvind Tafjord. 2018.
\newblock Think you have solved question answering? try arc, the ai2 reasoning
  challenge.
\newblock \emph{arXiv preprint arXiv:1803.05457}.

\bibitem[{Conneau et~al.(2017)Conneau, Lample, Ranzato, Denoyer, and
  J{\'e}gou}]{conneau2017word}
Alexis Conneau, Guillaume Lample, Marc'Aurelio Ranzato, Ludovic Denoyer, and
  Herv{\'e} J{\'e}gou. 2017.
\newblock Word translation without parallel data.
\newblock \emph{arXiv preprint arXiv:1710.04087}.

\bibitem[{Devlin et~al.(2018)Devlin, Chang, Lee, and
  Toutanova}]{devlin2018bert}
Jacob Devlin, Ming-Wei Chang, Kenton Lee, and Kristina Toutanova. 2018.
\newblock Bert: Pre-training of deep bidirectional transformers for language
  understanding.
\newblock \emph{arXiv preprint arXiv:1810.04805}.

\bibitem[{Ganin et~al.(2016)Ganin, Ustinova, Ajakan, Germain, Larochelle,
  Laviolette, Marchand, and Lempitsky}]{ganin2016domain}
Yaroslav Ganin, Evgeniya Ustinova, Hana Ajakan, Pascal Germain, Hugo
  Larochelle, Fran{\c{c}}ois Laviolette, Mario Marchand, and Victor Lempitsky.
  2016.
\newblock Domain-adversarial training of neural networks.
\newblock \emph{The Journal of Machine Learning Research}, 17(1):2096--2030.

\bibitem[{Garc{\'\i}a et~al.(2012)Garc{\'\i}a, Hurtado, Segarra, Sanchis, and
  Riccardi}]{garcia2012combining}
Fernando Garc{\'\i}a, Llu{\'\i}s~F Hurtado, Encarna Segarra, Emilio Sanchis,
  and Giuseppe Riccardi. 2012.
\newblock Combining multiple translation systems for spoken language
  understanding portability.
\newblock In \emph{Spoken Language Technology Workshop (SLT), 2012 IEEE}, pages
  194--198. IEEE.

\bibitem[{Golub et~al.(2017)Golub, Huang, He, and Deng}]{golub2017two}
David Golub, Po-Sen Huang, Xiaodong He, and Li~Deng. 2017.
\newblock Two-stage synthesis networks for transfer learning in machine
  comprehension.
\newblock \emph{arXiv preprint arXiv:1706.09789}.

\bibitem[{Gulrajani et~al.(2017)Gulrajani, Ahmed, Arjovsky, Dumoulin, and
  Courville}]{gulrajani2017improved}
Ishaan Gulrajani, Faruk Ahmed, Martin Arjovsky, Vincent Dumoulin, and Aaron~C
  Courville. 2017.
\newblock Improved training of wasserstein gans.
\newblock In \emph{Advances in Neural Information Processing Systems}, pages
  5767--5777.

\bibitem[{He et~al.(2017)He, Liu, Liu, Lyu, Zhao, Xiao, Liu, Wang, Wu, She
  et~al.}]{he2017dureader}
Wei He, Kai Liu, Jing Liu, Yajuan Lyu, Shiqi Zhao, Xinyan Xiao, Yuan Liu,
  Yizhong Wang, Hua Wu, Qiaoqiao She, et~al. 2017.
\newblock Dureader: a chinese machine reading comprehension dataset from
  real-world applications.
\newblock \emph{arXiv preprint arXiv:1711.05073}.

\bibitem[{He et~al.(2013)He, Deng, Hakkani-Tur, and Tur}]{he2013multi}
Xiaodong He, Li~Deng, Dilek Hakkani-Tur, and Gokhan Tur. 2013.
\newblock Multi-style adaptive training for robust cross-lingual spoken
  language understanding.
\newblock In \emph{Acoustics, Speech and Signal Processing (ICASSP), 2013 IEEE
  International Conference on}, pages 8342--8346. IEEE.

\bibitem[{Hu et~al.(2017)Hu, Peng, Huang, Qiu, Wei, and
  Zhou}]{hu2017reinforced}
Minghao Hu, Yuxing Peng, Zhen Huang, Xipeng Qiu, Furu Wei, and Ming Zhou. 2017.
\newblock Reinforced mnemonic reader for machine reading comprehension.
\newblock \emph{arXiv preprint arXiv:1705.02798}.

\bibitem[{Huang et~al.(2017)Huang, Zhu, Shen, and Chen}]{huang2017fusionnet}
Hsin-Yuan Huang, Chenguang Zhu, Yelong Shen, and Weizhu Chen. 2017.
\newblock Fusionnet: Fusing via fully-aware attention with application to
  machine comprehension.
\newblock \emph{arXiv preprint arXiv:1711.07341}.

\bibitem[{Joshi et~al.(2017)Joshi, Choi, Weld, and
  Zettlemoyer}]{joshi2017triviaqa}
Mandar Joshi, Eunsol Choi, Daniel~S Weld, and Luke Zettlemoyer. 2017.
\newblock Triviaqa: A large scale distantly supervised challenge dataset for
  reading comprehension.
\newblock \emph{arXiv preprint arXiv:1705.03551}.

\bibitem[{Kadlec et~al.(2016)Kadlec, Bajgar, and
  Kleindienst}]{kadlec2016particular}
Rudolf Kadlec, Ondrej Bajgar, and Jan Kleindienst. 2016.
\newblock From particular to general: A preliminary case study of transfer
  learning in reading comprehension.
\newblock In \emph{Machine Intelligence Workshop, NIPS}.

\bibitem[{Kim et~al.(2017)Kim, Kim, Sarikaya, and
  Fosler-Lussier}]{kim2017cross}
Joo-Kyung Kim, Young-Bum Kim, Ruhi Sarikaya, and Eric Fosler-Lussier. 2017.
\newblock Cross-lingual transfer learning for pos tagging without cross-lingual
  resources.
\newblock In \emph{Proceedings of the 2017 Conference on Empirical Methods in
  Natural Language Processing}, pages 2832--2838.

\bibitem[{Lai et~al.(2017)Lai, Xie, Liu, Yang, and Hovy}]{lai2017race}
Guokun Lai, Qizhe Xie, Hanxiao Liu, Yiming Yang, and Eduard Hovy. 2017.
\newblock Race: Large-scale reading comprehension dataset from examinations.
\newblock \emph{arXiv preprint arXiv:1704.04683}.

\bibitem[{Lample and Conneau(2019)}]{lample2019cross}
Guillaume Lample and Alexis Conneau. 2019.
\newblock Cross-lingual language model pretraining.
\newblock \emph{arXiv preprint arXiv:1901.07291}.

\bibitem[{Lee et~al.(2018)Lee, Wang, Chang, and Lee}]{lee2018odsqa}
Chia-Hsuan Lee, Shang-Ming Wang, Huan-Cheng Chang, and Hung-Yi Lee. 2018.
\newblock Odsqa: Open-domain spoken question answering dataset.
\newblock In \emph{2018 IEEE Spoken Language Technology Workshop (SLT)}, pages
  949--956. IEEE.

\bibitem[{Li et~al.(2018)Li, Wu, Liu, and Lee}]{li2018spoken}
Chia-Hsuan Li, Szu-Lin Wu, Chi-Liang Liu, and Hung-yi Lee. 2018.
\newblock Spoken squad: A study of mitigating the impact of speech recognition
  errors on listening comprehension.
\newblock \emph{arXiv preprint arXiv:1804.00320}.

\bibitem[{Liu et~al.(2017)Liu, Shen, Duh, and Gao}]{liu2017stochastic}
Xiaodong Liu, Yelong Shen, Kevin Duh, and Jianfeng Gao. 2017.
\newblock Stochastic answer networks for machine reading comprehension.
\newblock \emph{arXiv preprint arXiv:1712.03556}.

\bibitem[{Lu et~al.(2011)Lu, Tan, Cardie, and Tsou}]{lu2011joint}
Bin Lu, Chenhao Tan, Claire Cardie, and Benjamin~K Tsou. 2011.
\newblock Joint bilingual sentiment classification with unlabeled parallel
  corpora.
\newblock In \emph{Proceedings of the 49th Annual Meeting of the Association
  for Computational Linguistics: Human Language Technologies-Volume 1}, pages
  320--330. Association for Computational Linguistics.

\bibitem[{Magnini et~al.(2006)Magnini, Giampiccolo, Forner, Ayache, Jijkoun,
  Osenova, Pe{\~n}as, Rocha, Sacaleanu, and Sutcliffe}]{magnini2006overview}
Bernardo Magnini, Danilo Giampiccolo, Pamela Forner, Christelle Ayache,
  Valentin Jijkoun, Petya Osenova, Anselmo Pe{\~n}as, Paulo Rocha, Bogdan
  Sacaleanu, and Richard Sutcliffe. 2006.
\newblock Overview of the clef 2006 multilingual question answering track.
\newblock In \emph{Workshop of the Cross-Language Evaluation Forum for European
  Languages}, pages 223--256. Springer.

\bibitem[{McClosky et~al.(2010)McClosky, Charniak, and
  Johnson}]{mcclosky2010automatic}
David McClosky, Eugene Charniak, and Mark Johnson. 2010.
\newblock Automatic domain adaptation for parsing.
\newblock In \emph{Human Language Technologies: The 2010 Annual Conference of
  the North American Chapter of the Association for Computational Linguistics},
  pages 28--36. Association for Computational Linguistics.

\bibitem[{Min et~al.(2017)Min, Seo, and Hajishirzi}]{min2017question}
Sewon Min, Minjoon Seo, and Hannaneh Hajishirzi. 2017.
\newblock Question answering through transfer learning from large fine-grained
  supervision data.
\newblock \emph{arXiv preprint arXiv:1702.02171}.

\bibitem[{Mohammad et~al.(2016)Mohammad, Salameh, and
  Kiritchenko}]{mohammad2016translation}
Saif~M Mohammad, Mohammad Salameh, and Svetlana Kiritchenko. 2016.
\newblock How translation alters sentiment.
\newblock \emph{Journal of Artificial Intelligence Research}, 55:95--130.

\bibitem[{Mori and Takahashi(2007)}]{mori2007method}
Tatsunori Mori and Kousuke Takahashi. 2007.
\newblock A method of cross-lingual question-answering based on machine
  translation and noun phrase translation using web documents.
\newblock In \emph{Proceedings of NTCIR-6 Workshop, Tokyo, Japan}.

\bibitem[{Mulcaire et~al.(2019)Mulcaire, Kasai, and
  Smith}]{mulcaire2019polyglot}
Phoebe Mulcaire, Jungo Kasai, and Noah Smith. 2019.
\newblock Polyglot contextual representations improve crosslingual transfer.
\newblock \emph{arXiv preprint arXiv:1902.09697}.

\bibitem[{Nguyen et~al.(2016)Nguyen, Rosenberg, Song, Gao, Tiwary, Majumder,
  and Deng}]{nguyen2016ms}
Tri Nguyen, Mir Rosenberg, Xia Song, Jianfeng Gao, Saurabh Tiwary, Rangan
  Majumder, and Li~Deng. 2016.
\newblock Ms marco: A human generated machine reading comprehension dataset.
\newblock \emph{arXiv preprint arXiv:1611.09268}.

\bibitem[{Rajpurkar et~al.(2016)Rajpurkar, Zhang, Lopyrev, and
  Liang}]{rajpurkar2016squad}
Pranav Rajpurkar, Jian Zhang, Konstantin Lopyrev, and Percy Liang. 2016.
\newblock Squad: 100,000+ questions for machine comprehension of text.
\newblock \emph{arXiv preprint arXiv:1606.05250}.

\bibitem[{Rohrbach et~al.(2015)Rohrbach, Rohrbach, Tandon, and
  Schiele}]{rohrbach2015dataset}
Anna Rohrbach, Marcus Rohrbach, Niket Tandon, and Bernt Schiele. 2015.
\newblock A dataset for movie description.
\newblock In \emph{Proceedings of the IEEE conference on computer vision and
  pattern recognition}, pages 3202--3212.

\bibitem[{Sasaki et~al.(2005)Sasaki, Chen, Chen, and Lin}]{sasaki2005overview}
Yutaka Sasaki, Hsin-Hsi Chen, Kuang-hua Chen, and Chuan-Jie Lin. 2005.
\newblock Overview of the ntcir-5 cross-lingual question answering task
  (clqa1).
\newblock In \emph{NTCIR}.

\bibitem[{Schuster et~al.(2018)Schuster, Gupta, Shah, and
  Lewis}]{schuster2018cross}
Sebastian Schuster, Sonal Gupta, Rushin Shah, and Mike Lewis. 2018.
\newblock Cross-lingual transfer learning for multilingual task oriented
  dialog.
\newblock \emph{arXiv preprint arXiv:1810.13327}.

\bibitem[{Seo et~al.(2016)Seo, Kembhavi, Farhadi, and
  Hajishirzi}]{seo2016bidirectional}
Minjoon Seo, Aniruddha Kembhavi, Ali Farhadi, and Hannaneh Hajishirzi. 2016.
\newblock Bidirectional attention flow for machine comprehension.
\newblock \emph{arXiv preprint arXiv:1611.01603}.

\bibitem[{Shao et~al.(2018)Shao, Liu, Lai, Tseng, and Tsai}]{shao2018drcd}
Chih~Chieh Shao, Trois Liu, Yuting Lai, Yiying Tseng, and Sam Tsai. 2018.
\newblock Drcd: a chinese machine reading comprehension dataset.
\newblock \emph{arXiv preprint arXiv:1806.00920}.

\bibitem[{Shinohara(2016)}]{shinohara2016adversarial}
Yusuke Shinohara. 2016.
\newblock Adversarial multi-task learning of deep neural networks for robust
  speech recognition.
\newblock In \emph{INTERSPEECH}, pages 2369--2372.

\bibitem[{Stepanov et~al.(2013)Stepanov, Kashkarev, Bayer, Riccardi, and
  Ghosh}]{stepanov2013language}
Evgeny~A Stepanov, Ilya Kashkarev, Ali~Orkan Bayer, Giuseppe Riccardi, and
  Arindam Ghosh. 2013.
\newblock Language style and domain adaptation for cross-language slu porting.
\newblock In \emph{Automatic Speech Recognition and Understanding (ASRU), 2013
  IEEE Workshop on}, pages 144--149. IEEE.

\bibitem[{Trischler et~al.(2016)Trischler, Wang, Yuan, Harris, Sordoni,
  Bachman, and Suleman}]{trischler2016newsqa}
Adam Trischler, Tong Wang, Xingdi Yuan, Justin Harris, Alessandro Sordoni,
  Philip Bachman, and Kaheer Suleman. 2016.
\newblock Newsqa: A machine comprehension dataset.
\newblock \emph{arXiv preprint arXiv:1611.09830}.

\bibitem[{Wang et~al.(2018)Wang, Yan, and Wu}]{wang2018multi}
Wei Wang, Ming Yan, and Chen Wu. 2018.
\newblock Multi-granularity hierarchical attention fusion networks for reading
  comprehension and question answering.
\newblock \emph{arXiv preprint arXiv:1811.11934}.

\bibitem[{Wang et~al.(2017)Wang, Yang, Wei, Chang, and Zhou}]{wang2017gated}
Wenhui Wang, Nan Yang, Furu Wei, Baobao Chang, and Ming Zhou. 2017.
\newblock Gated self-matching networks for reading comprehension and question
  answering.
\newblock In \emph{Proceedings of the 55th Annual Meeting of the Association
  for Computational Linguistics (Volume 1: Long Papers)}, volume~1, pages
  189--198.

\bibitem[{Wiese et~al.(2017)Wiese, Weissenborn, and Neves}]{wiese2017neural}
Georg Wiese, Dirk Weissenborn, and Mariana Neves. 2017.
\newblock Neural domain adaptation for biomedical question answering.
\newblock \emph{arXiv preprint arXiv:1706.03610}.

\bibitem[{Xiong et~al.(2016)Xiong, Zhong, and Socher}]{xiong2016dynamic}
Caiming Xiong, Victor Zhong, and Richard Socher. 2016.
\newblock Dynamic coattention networks for question answering.
\newblock \emph{arXiv preprint arXiv:1611.01604}.

\bibitem[{Yang et~al.(2017)Yang, Salakhutdinov, and Cohen}]{yang2017transfer}
Zhilin Yang, Ruslan Salakhutdinov, and William~W Cohen. 2017.
\newblock Transfer learning for sequence tagging with hierarchical recurrent
  networks.
\newblock \emph{arXiv preprint arXiv:1703.06345}.

\bibitem[{Yu et~al.(2018)Yu, Dohan, Luong, Zhao, Chen, Norouzi, and
  Le}]{yu2018qanet}
Adams~Wei Yu, David Dohan, Minh-Thang Luong, Rui Zhao, Kai Chen, Mohammad
  Norouzi, and Quoc~V Le. 2018.
\newblock Qanet: Combining local convolution with global self-attention for
  reading comprehension.
\newblock \emph{arXiv preprint arXiv:1804.09541}.

\bibitem[{Zirikly and Hagiwara(2015)}]{zirikly2015cross}
Ayah Zirikly and Masato Hagiwara. 2015.
\newblock Cross-lingual transfer of named entity recognizers without parallel
  corpora.
\newblock In \emph{Proceedings of the 53rd Annual Meeting of the Association
  for Computational Linguistics and the 7th International Joint Conference on
  Natural Language Processing (Volume 2: Short Papers)}, volume~2, pages
  390--396.

\bibitem[{Zitnick et~al.(2016)Zitnick, Vedantam, and
  Parikh}]{zitnick2016adopting}
C~Lawrence Zitnick, Ramakrishna Vedantam, and Devi Parikh. 2016.
\newblock Adopting abstract images for semantic scene understanding.
\newblock \emph{IEEE transactions on pattern analysis and machine
  intelligence}, 38(4):627--638.

\end{thebibliography}
\bibliographystyle{acl_natbib}

\appendix

\section{Supplemental Material to accompany \emph{Cross-Lingual Transfer Learning for Question Answering}}
\label{sec:supplemental}
\subsection{Test-on-Source}
In the EQA task, the ground-truth answer is always a span in the document.
However, in the test-on-source approach, the testing data is transcribed into English.
It is possible that after translation, the translation of answer is not in the translation of its corresponding document; therefore we compiled DRCD$_{\mathit{filter}}$, a filtered testing set from DRCD, in which only the examples fulfilling the requirement of EQA after translation are included. We report the performance on both the original DRCD testing set and the DRCD$_{\mathit{filter}}$ testing set.

For the test-on-source approach, all the Chinese examples in DRCD, including the training and testing sets, were transcribed into English and they are denoted as DRCD (English). All results of the various test-on-source settings are reported in Table~\ref{tab:testonsource-supplement}.

We report the evaluation results on both DRCD$_{\mathit{filter}}$ (English) and DRCD (English). 
Because the testing documents were translated into English, the output predictions of QA model were also English (column (1), column (3)).
For evaluation, all the ground-truth answers in this case were translated from the original Chinese answers using Google Machine Translation.
We also back-translated the English predictions into Chinese (column (2), column (4)), and evaluated the performance.

As you can see in Table~\ref{tab:testonsource-supplement}, the EM and F1 scores on Chinese answers (column (2), column (4)) are much lower than that for the English answers (column (1), column (3)).  
This was expected because the Chinese predictions were translated from the English predictions; the resulting translation errors degraded performance. 
However, when comparing the different approaches, the conclusion drawn from the Chinese and English evaluations is the same:
They both show that training from DRCD (English) outperforms SQuAD even though SQuAD has more training examples.
This is intuitive because the data distribution of DRCD (English) is closer to the testing data here. 
Using both corpora is even better (row (e) vs. row (d)).

\subsection{Train-on-Target}
We also report the performance of Train-on-Target on DRCD$_{\mathit{filter}}$ in Table~\ref{tab:MT-supplement}.

\subsection{GAN-based Approach}
We also report the performance of GAN-based approaches on DRCD$_{\mathit{filter}}$ in Table~\ref{tab:GAN-supplement}.

\begin{table}[ht]
\centering
\caption{EM/F1 train-on-target scores over DRCD$_{\mathit{filter}}$ and DRCD. SQuAD-zh denotes SQuAD (MT) and NewsQA-zh denotes NewsQA (MT)}
\vspace*{-2mm}
\label{tab:MT-supplement}
\begin{tabular}{|c|c|c|c|c|c|c|c|}
\hline
\multicolumn{2}{|c|}{{}} &
\multicolumn{1}{|c|}{{}}&
\multicolumn{2}{|c|}{{DRCD$_{\mathit{filter}}$}} & \multicolumn{2}{|c|}{{DRCD}} \\
\cline{4-7}
\multicolumn{2}{|c|}{{Approaches}} & \multicolumn{1}{|c|}{} & EM & F1 & EM & F1   \\
\hline
\hline
\multicolumn{2}{|c|}{{\cite{shao2018drcd}}} & (a) & - & - & - & 53.78 \\
\multicolumn{2}{|c|}{{Baseline}} & (b)  & 67.72 & 79.23 & 66.10 & 78.01 \\
\multicolumn{2}{|c|}{{Human}} & (c) &- & - & 80.43 & 93.30 \\
\hline\hline
\multicolumn{2}{|c|}{{\emph{SQuAD-zh}}} & (d) & 59.00 & 76.21 & 53.50 & 72.22 \\
\multicolumn{2}{|c|}{{+DRCD}} & (e) & 77.61 & 87.62 & 74.20 & 85.67 \\
\hline\hline
\multicolumn{2}{|c|}{\emph{{NewsQA-zh}}} & (g) & 24.76 & 38.7 & 22.42 & 35.99 \\
\multicolumn{2}{|c|}{{+DRCD}} & (h) & 70.49 & 83.53 & 68.98 & 81.41 \\
\hline
\end{tabular}
\end{table}

\begin{table}[ht]
\centering
\caption{EM/F1 scores of GAN-based approaches over DRCD$_{\mathit{filter}}$ and DRCD.}
\vspace*{-2mm}
\label{tab:GAN-supplement}
\begin{tabular}{|c|c|c|c|c|c|c|}
\hline
\multicolumn{2}{|c|}{{}} &
\multicolumn{1}{|c|}{{}}&
\multicolumn{2}{|c|}{{DRCD$_{\mathit{filter}}$}} &
\multicolumn{2}{|c|}{{DRCD}} \\
\hline
\multicolumn{2}{|c|}{{Approaches}} & \multicolumn{1}{|c|}{} & EM & F1 & EM & F1 \\
\hline
\hline
\multicolumn{2}{|c|}{{Baseline}} & (a) & 67.72 & 79.23 & 66.10 & 78.01 \\
\hline\hline
\multicolumn{7}{|c|}{{\emph{SQuAD}}}\\
\hline
\multicolumn{2}{|c|}{{MT}}  & (b) & 77.61 & 87.62 & 74.20 & 85.67 \\
\multicolumn{2}{|c|}{{Dependent}} & (c)& 73.18 & 83.10 & 70.97 & 81.92 \\
\multicolumn{2}{|c|}{{+GAN-CH}} & (d) & 77.04 & 87.51 & 74.00 &  85.35\\
\multicolumn{2}{|c|}{{+GAN-EN}} & (e) & 72.45 & 85.27 & 70.97 & 83.36 \\
\multicolumn{2}{|c|}{{MT+GAN-CH}} & (f) & 78.21 & 89.36 & 75.12 & 87.26 \\
\hline\hline
\multicolumn{7}{|c|}{{\emph{NewsQA}}}\\
\hline
\multicolumn{2}{|c|}{{MT}}  & (g) & 70.49 & 83.53 & 68.98 & 81.41 \\
\multicolumn{2}{|c|}{{Dependent}} & (h)& 71.20 & 82.12 & 67.96 & 80.51 \\
\multicolumn{2}{|c|}{{+GAN-CH}} & (i) & 75.54 & 86.35 & 71.73 & 83.90\\
\multicolumn{2}{|c|}{{+GAN-EN}} & (j) & 72.58 & 84.36 & 69.25 & 82.28 \\
\multicolumn{2}{|c|}{{MT+GAN-CH}} & (k) & 76.83 & 87.22 & 72.79 & 84.94 \\
\hline
\end{tabular}
\end{table}

\begin{table*}[b]
\centering
\caption{EM/F1 test-on-source scores over DRCD$_{\mathit{filter}}$ (English) and DRCD (English).
Because the testing documents are translated into English, the output predictions of the QA model are also in English (column (1), column (3)).
We also back-translated the English predictions into Chinese (column (2), column (4)) and evaluated with ground-truths from DRCD$_{\mathit{filter}}$ and DRCD respectively.}
\label{tab:testonsource-supplement}
\begin{tabular}{|c|c|c|c|c|c|c|c|c|c|c|}
\hline
\multicolumn{2}{|c|}{{}} &
\multicolumn{1}{|c|}{{}}&
\multicolumn{4}{|c|}{{DRCD$_{\mathit{filter}}$ (English)}} &
\multicolumn{4}{|c|}{{DRCD (English) }}  \\
\cline{4-11}
\multicolumn{2}{|c|}{{}} &
\multicolumn{1}{|c|}{{}}&
\multicolumn{2}{|c|}{{(1) ENG Ans}} & \multicolumn{2}{|c|}{{(2) CH Ans }}&
\multicolumn{2}{|c|}{{(3) ENG Ans }} & \multicolumn{2}{|c|}{{(4) CH Ans }} \\
\cline{4-11}
\multicolumn{2}{|c|}{{Approaches}} & \multicolumn{1}{|c|}{} & EM & F1 & EM & F1 & EM & F1 & EM & F1   \\
\hline
\hline
\multicolumn{2}{|c|}{{\cite{shao2018drcd}}} & (a) & - & - & - & - & - & - & - & 53.78\\
\multicolumn{2}{|c|}{{Baseline}} & (b) & - & -  & 67.72 & 79.23 & - & - & 66.10 & 78.01\\
\hline\hline
\multicolumn{2}{|c|}{{SQuAD}}    & (c)  & 49.75 & 60.61 & 30.81 & 51.91 & 34.67 & 49.31 & 23.43 & 45.29 \\
\multicolumn{2}{|c|}{{DRCD (English)}}    & (d)   &  53.97  & 63.10 & 32.40 & 53.85 & 38.02 & 51.30 & 24.57 & 46.07\\
\multicolumn{2}{|c|}{{SQuAD + DRCD (English)}}    & (e)   &  59.44  & 68.72 & 35.60 & 58.03 & 42.02 & 56.23  & 27.09 & 50.44 \\
\hline
\end{tabular}
\end{table*}

\end{document}